\documentclass[11pt]{article}

\usepackage[preprint]{acl}

\usepackage{times}
\usepackage{latexsym}

\usepackage[T1]{fontenc}
\usepackage[utf8]{inputenc}

\usepackage{microtype}
\IfFileExists{inconsolata.sty}{\usepackage{inconsolata}}{}
\usepackage{graphicx}

\usepackage{booktabs}
\usepackage{amsmath}
\usepackage{amssymb}
\usepackage{multirow}
\usepackage{listings}
\usepackage{xcolor}
\usepackage{algorithm}
\usepackage{algpseudocode}
\algrenewcommand{\algorithmiccomment}[1]{\hfill{\textcolor{gray}{\scriptsize // #1}}}
\usepackage{tikz}
\usetikzlibrary{positioning,arrows.meta,shapes.geometric,calc,fit,backgrounds}
\usepackage{pgfplots}
\usepackage{subcaption}
\pgfplotsset{compat=1.18}

\newcommand{\jinasmall}{\texttt{j-v5-small}}
\newcommand{\jinanano}{\texttt{j-v5-nano}}
\newcommand{\efivebase}{\texttt{e5-base-v2}}
\newcommand{\efivelarge}{\texttt{e5-large-v2}}
\newcommand{\efivesmall}{\texttt{e5-small-v2}}
\newcommand{\gtebase}{\texttt{gte-base-en-v1.5}}

\newcommand{\bgelarge}{\texttt{bge-large-en-v1.5}}

\newcommand{\jinafamily}{\texttt{j-v5}}
\newcommand{\qwenthreeemb}{\texttt{qwen3-0.6b}}
\newcommand{\embgemma}{\texttt{gemma-300m}}

\title{Test-Time Compute for Frozen Embedding Models through Agentic Program Search}

\author{Han Xiao \\
  Jina AI \emph{by} Elastic \\
  \texttt{han.xiao@jina.ai}}

\begin{document}
\maketitle

\begin{abstract}
Test-time compute is widely believed to benefit only large reasoning models. We argue the opposite for dense retrieval, since modern small embedding models are distilled or adapted from large language model backbones and can inherit their test-time-compute potential. We ask how much retrieval quality a frozen embedding model gains at inference alone, with no auxiliary model and no parameters trained at deployment. An agentic loop in which a large language model writes programs over a frozen encoder API yields twelve Pareto-optimal programs that trade inference compute for quality across cost ratios from $c{=}1.2$ to $14.7$, every one with a positive mean $\Delta$nDCG@10 across the 14 discovery tasks, while recovering classical primitives such as reciprocal rank fusion, the Fisher linear discriminant, Rocchio feedback, and MaxSim. Applied unmodified to nineteen held-out tasks and three unseen encoder families, a single fixed program improves the majority of cells, with a positive median $\Delta$nDCG@10 and a $54\%$ win-rate, and gains most on the unseen families. A matched-budget learned projection head trained on the same tasks instead overfits the discovery domains and falls below baseline on every held-out encoder. Small embedding models therefore inherit usable test-time-compute potential that transfers to new corpora and encoders with no per-domain labels.
\end{abstract}

\section{Introduction}
\label{sec:intro}

Test-time compute has transformed reasoning in generative language models. Best-of-$N$ sampling \citep{brown2024monkeys}, self-consistency \citep{wang2023selfconsistency}, and verifier-guided search \citep{snell2024scalingttc,wu2024inferencescaling} let a smaller model spend more inference FLOPs to match a larger one along a smooth Pareto curve. The prevailing assumption is that this scaling regime is exclusive to large reasoning LLMs and that small models have nothing to gain. Dense retrieval, dominated by small embedding models, has not yet been studied through a test-time-compute lens. Frozen embedding models such as \texttt{jina-embeddings-v5-text-small} (\jinasmall{}), \efivebase{}, and \gtebase{} are deployed with one forward pass per query and one cosine similarity per document.

We argue that the small-model exclusion is misleading, because most modern embedding models are distilled or adapted from large LLM backbones and inherit their representational capacity. E5-mistral first turned Mistral-7B into a state-of-the-art embedder from synthetic data alone \citep{wang2024e5mistral}, and the recipe spread across SFR-Embedding \citep{meng2024sfr}, GritLM \citep{muennighoff2024gritlm}, NV-Embed \citep{lee2024nvembed}, and bge-en-icl \citep{li2024bgeenicl} on Mistral-7B, and RepLLaMA on LLaMA \citep{ma2024repllama}. The pattern spans backbone families, including \qwenthreeemb{} from Qwen3 \citep{zhang2025qwen3embedding}, \embgemma{} from Gemma~3 \citep{vera2025embeddinggemma}, and \jinafamily{} distilled from Qwen3 with task-LoRA adapters \citep{akram2026jinav5}. To the extent that test-time compute is a property of the underlying LLM representation space, these distilled small embedding models should inherit at least some of that potential. Existing test-time scaling proposals for retrieval instead require an external generative LLM \citep{gao2023hyde,wang2023query2doc}, a second supervisory retriever \citep{uzan2025gqr}, or trained extra parameters \citep{xiao2025metaembed}. We therefore ask a stricter question. How much can a frozen single-vector embedding model gain at inference alone, with no auxiliary model and no learned parameters?

We answer with an agentic program-search loop in which an LLM agent writes Python programs over the frozen embedding API, a harness scores each candidate on a multi-task retrieval benchmark, and a long-horizon memory accumulates ruled-out hypotheses. Across 144 generations, the loop produces twelve Pareto-optimal programs that improve retrieval quality at cost ratios from $c{=}1.2$ to $14.7$. We validate generalization by applying the discovered programs, without modification, to held-out encoder families and retrieval tasks not seen during discovery.

Our study addresses how the programs are generated and evolve, how their inference cost trades for retrieval quality, whether a program learned on a small discovery set keeps its gain on full held-out tasks and on unseen encoder families, and what the universally useful programs look like relative to classical retrieval. We further compare this inference-time axis against the training-time axis it is meant to substitute for, since the canonical scaling results trade inference compute against training compute along a Pareto curve \citep{snell2024scalingttc}. A matched-budget projection head trained on the same discovery tasks does not generalize the way these programs do, as Appendix~\ref{app:headbaseline} shows, which indicates that the transfer is a property of inference-time structure rather than of any learned metric. The program search itself consumes labeled supervision on the 14 discovery tasks, so the method is label-free at deployment rather than end-to-end, and what transfers without further labels is the fixed program it produces.

\section{Related Work}
\label{sec:related}

\subsection{Test-time compute for generative LLMs}
Best-of-$N$ scaling forms the dominant test-time compute paradigm for generative LLMs. \citet{snell2024scalingttc} formalize compute-optimal allocation between best-of-$N$ sampling and iterative revision. \citet{brown2024monkeys} show best-of-$N$ scales coverage log-linearly. \citet{wu2024inferencescaling} establish a clean inference scaling Pareto frontier on math problem-solving. \citet{wang2023selfconsistency} introduce majority-vote over sampled chains-of-thought. Common to all four is stochasticity: sample many noisy candidates, then aggregate. We adopt the same question: whether extra inference compute can substitute for a larger embedding model, and find the embedding analogue is structural rather than stochastic.

\subsection{Generation-based query expansion}
HyDE \citep{gao2023hyde} embeds an LLM-generated hypothetical answer in place of the query; Query2Doc \citep{wang2023query2doc} concatenates the LLM expansion to the query before encoding. Both rely on an external LLM at query time, trading inference latency and cost for retrieval gain. \citet{zhuang2024promptreps} prompt LLMs to generate dense+sparse representations directly. We restrict ourselves to the embedding model itself, with no generative model in the inference path.

\subsection{Pseudo-relevance feedback}
Classical PRF \citep{rocchio1971relevance,robertson1995okapi} re-formulates the query using terms from initial top-ranked documents. The de-facto sparse-PRF baseline is RM3 \citep{lavrenko2001rm3}, which interpolates the original query with a relevance language model estimated from pseudo-relevant documents and remains a strong baseline on top of BM25. Learned sparse retrievers such as SPLADE \citep{formal2021splade} likewise rely on lexical expansion.

Recent dense PRF work takes three distinct routes. CEQE \citep{naseri2021ceqe} operates at the term level via contextualized embeddings. VPRF \citep{li2023vprf} aggregates top-$k$ dense embeddings into a vector-PRF query update. ANCE-PRF \citep{yu2021anceprf,li2022anceprfrepro} trains a small reranker to consume top-$k$ embeddings. VPRF is the closest training-free prior art, since it computes a uniform mean over the top-$k$ retrieved-document embeddings and interpolates with the query. A separate line of recent vector-PRF extensions brings an LLM into the inference loop, including LLM-VPRF \citep{li2025llmvprf}, GPRF \citep{tu2025gprf}, and PromptPRF \citep{li2025promptprf}. These lie in a different inference regime than the training-free, no-LLM substrate explored here. PromptPRF reports that LLM-extracted features over PRF documents narrow the gap between small and large dense retrievers. We observe a similar small-model lift without requiring an LLM in the inference path.

The difference from these feedback baselines is one of inference-time cost rather than of mechanism. ColBERT-PRF extends MaxSim feedback into multi-vector space at a per-token storage and late-interaction cost \citep{wang2021colbertprf}, whereas our sentence-level MaxSim stays within the single-vector framework, and the generative-feedback methods, including generative relevance feedback \citep{mackie2023grfprf}, incur either an online generation call per query or an offline index pass that re-encodes documents with LLM-derived features. Our substrate adds no generative model and no learned parameters at inference, so rather than running these methods as primary baselines we isolate the shared feedback mechanism through the controlled \textsc{SoftCentroid}-versus-Rocchio/VPRF comparison of Appendix~\ref{app:rocchio-comparison}, and we leave fuller shared-matrix comparisons against the LLM-driven and multi-vector variants as a limitation.

Our agentic search arrives at Rocchio PRF after being given it as one of several research inspirations (Section~\ref{sec:proposer}); the appearance of PRF is thus an operationalization of a seeded idea rather than an unprompted rediscovery, and it confirms that dense PRF remains a robust mechanism in modern embedding spaces. The unprompted rediscoveries, not present in the inspiration set, are reciprocal rank fusion and the Fisher linear discriminant.

\subsection{Test-time scaling for retrieval}
MetaEmbed \citep{xiao2025metaembed} adds learnable Meta-Tokens that produce a flexible multi-vector representation, scaling accuracy with the number of tokens kept at test time. It requires training. GQR \citep{uzan2025gqr} performs gradient descent on the query embedding using a second model's similarity as supervision. It requires a secondary retriever. Both target multimodal hybrid retrieval. Late interaction in ColBERT and ColBERTv2 \citep{khattab2020colbert,santhanam2022colbertv2} is another way to spend more compute per query, but at the cost of multi-vector storage. Our regime is strictly more constrained, requiring a frozen single-vector embedding model, no second model, and no extra forward pass for the cheapest variant.

\subsection{Program generation and agentic discovery}
Closest to our methodological framing are LLM-driven program-generation loops in which a language model proposes candidate programs, an evaluator scores them, and the surviving programs condition the next round of proposals. FunSearch \citep{romera2024funsearch} discovered new constructions for the cap set problem and bin packing this way. AlphaEvolve \citep{novikov2025alphaevolve} extended the pattern to a coding agent capable of improving algorithmic primitives. ELM \citep{lehman2023elm} explored LLM-assisted evolution earlier. Our agent loop is an instance of the same pattern, specialized to embedding-program search. The agent is conditioned on a running frontier of non-dominated programs over cost and $\Delta$nDCG, plus a structured log of which substrate families the search has already moved beyond. The evaluator is a pinned multi-task retrieval harness. To our knowledge this is the first time this loop has been applied to the dense-retrieval inference recipe.

\section{Agentic Program Discovery}
\label{sec:meta}

\begin{figure*}[t]
\centering
\begin{tikzpicture}[
  font=\footnotesize, >={Latex[length=1.8mm]},
  box/.style={rounded corners=2pt, draw=black!55, fill=blue!5, align=center,
              inner sep=3pt, minimum height=8mm, text width=26mm},
  ar/.style={-{Latex[length=1.8mm]}, draw=black!65, semithick},
  lbl/.style={font=\scriptsize\itshape, text=black!55, align=center},
  panel/.style={rounded corners=3pt, draw=black!25, fill=black!3, inner sep=3.2mm},
]
\node[box] (enc) {Frozen encoder $f_\theta$\\ multi-LoRA adapters};
\node[box, below=4mm of enc] (prog) {Program $P$\\ chunk, transform,\\ re-embed via \texttt{encode\_fn}};
\node[box, below=4mm of prog] (score) {Similarity scores $\mathbf{S}'$};
\node[lbl, below=2mm of score, text width=30mm] (con) {training-free,\\ six constraints};
\draw[ar] (enc) -- (prog);
\draw[ar] (prog) -- (score);
\begin{scope}[on background layer]
  \node[panel, fit=(enc)(con)] (pA) {};
\end{scope}
\node[anchor=south] at (pA.north) {\textbf{(a)} search space};

\node[box, right=11mm of enc] (prop) {Proposer\\ (Claude Opus)};
\node[box, right=10mm of prop] (eval) {Evaluator\\ $T{=}14$ tasks};
\node[box, below=12mm of eval] (front) {Frontier $\mathcal{F}$\\ cost vs.\ $\Delta$nDCG};
\node[box, below=12mm of prop] (mem) {Memory $\mathcal{H}$\\ JSONL ledger};
\draw[ar] (prop) -- (eval);
\draw[ar] (eval) -- (front);
\draw[ar] (front) -- (mem);
\draw[ar] (mem) -- (prop);
\node[lbl] at ($(prop)!0.5!(front)$) {$g=1\dots G$};
\begin{scope}[on background layer]
  \node[panel, fit=(prop)(eval)(front)(mem)] (pB) {};
\end{scope}
\node[anchor=south] at (pB.north) {\textbf{(b)} agentic search loop};

\node[box, right=11mm of eval] (reg) {Registry $\mathcal{R}$\\ 144 programs};
\node[box, below=4mm of reg] (par) {12 Pareto-optimal\\ $c{=}1.2$ to $14.7$};
\node[box, below=4mm of par, text width=30mm] (cls) {rediscovered: RRF, Fisher\\ operationalized: Rocchio, MaxSim};
\draw[ar] (reg) -- (par);
\draw[ar] (par) -- (cls);
\begin{scope}[on background layer]
  \node[panel, fit=(reg)(cls)] (pC) {};
\end{scope}
\node[anchor=south] at (pC.north) {\textbf{(c)} discovered programs};

\draw[ar] (prog.east) -- ([xshift=-0.5mm]pB.west|-prog);
\draw[ar] (eval.east) -- (reg.west);
\end{tikzpicture}
\caption{Overview of the agentic program discovery framework. \textbf{(a)}~The search space is defined by a frozen multi-LoRA encoder $f_\theta$ and a program interface: each program $P$ is an arbitrary Python function over \texttt{encode\_fn} subject to six constraints. \textbf{(b)}~The agentic loop runs for $G{=}144$ generations. At each round $g$, the proposer (Claude Opus~4.6) reads the frontier source code and structured history, then outputs a new program with a structural-novelty claim and hypothesis. The evaluator scores the program on $T{=}14$ retrieval tasks across legal, financial, long-document, and general domains, recording per-task $\Delta$nDCG@10 and cost ratio $c$. Surviving programs update the running frontier~$\mathcal{F}$; all results enter the JSONL memory~$\mathcal{H}$. \textbf{(c)}~After the loop terminates, the registry~$\mathcal{R}$ holds all 144 evaluated programs, of which 12 are identified as Pareto-optimal spanning $c{=}1.2$ to $14.7$. Several discovered programs converge to classical techniques such as Rocchio PRF, ColBERT MaxSim, reciprocal rank fusion, and Fisher linear discriminant analysis.}
\label{fig:architecture}
\end{figure*}
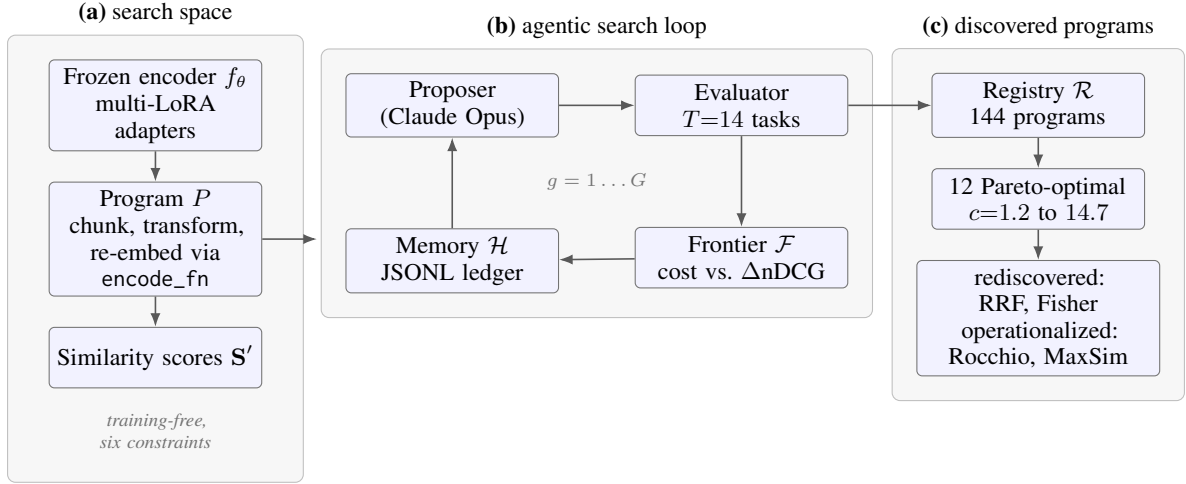

Our framework consists of five interacting modules (Figure~\ref{fig:architecture}).\footnote{Code, program registry, long-horizon memory, and evaluation harness: \url{https://github.com/hanxiao/embedding-ttc}} A frozen multi-LoRA encoder exposes a fixed API that programs call at inference time. A proposer (LLM agent) reads the current frontier and structured history, then writes a new Python program. An evaluator scores the program on 14 retrieval tasks and records per-task results and a cost ratio. Surviving programs update the frontier; all outcomes enter the memory. The loop repeats for $G$ generations, accumulating a registry of discovered programs. We describe each module below.

\subsection{Search space}
\label{sec:program-interface}

The search space is the program interface rather than a grammar, so that a restricted DSL cannot silently exclude the strategies we aim to discover. Each program $P$ is an arbitrary deterministic Python function
\[
  P: (\mathbf{Q}, \mathbf{D}, \mathbf{S}, \texttt{ctx}) \to \mathbf{S}' \in \mathbb{R}^{|Q|\times|D|}
\]
where $\mathbf{Q} \in \mathbb{R}^{|Q|\times d}$ and $\mathbf{D} \in \mathbb{R}^{|D|\times d}$ are L2-normalized query and document embeddings from a frozen encoder $f_\theta$, and $\mathbf{S} = \mathbf{Q}\mathbf{D}^\top$ is the baseline cosine similarity matrix. The context object \texttt{ctx} provides two capabilities. First, \texttt{encode\_fn} exposes the frozen encoder with selectable LoRA adapters for retrieval-query, retrieval-passage, classification, and text-matching, together with Matryoshka dimension truncation and input-length control. Each invocation of \texttt{encode\_fn} constitutes one unit of test-time compute. Second, \texttt{ctx} supplies raw query and document texts, corpus metadata, and identifier mappings.

The trivial program $P_0$ returns $\mathbf{S}$ unchanged and spends no test-time compute, and any \texttt{encode\_fn} call beyond it does. A multi-LoRA encoder exposes several task-specialized views of the same text for the agent to compose, and on encoders without adapters those channels collapse to a single view so that the structural operations alone carry the gain (Section~\ref{sec:xmodel-model}).

\subsection{Proposer}
\label{sec:proposer}

The proposer is Claude Opus~4.6 prompted to generate one Python program per generation. Its context window receives three inputs: a structured summary of the evaluation history $\mathcal{H}$ including per-task $\Delta$nDCG@10, win/tie/loss counts, and cost ratios for every previously evaluated program; the complete source code of the current top-$k$ frontier programs; and a set of research inspirations drawn from the retrieval literature, including Rocchio pseudo-relevance feedback, ColBERT-style MaxSim, cross-adapter voting, Matryoshka cascading, document substructure decomposition, and score-distribution analysis. Because Rocchio PRF and MaxSim are named in this inspiration set, their emergence in the frontier is the agent \emph{operationalizing} a seeded idea, not an unprompted rediscovery; we reserve the word ``rediscovery'' for reciprocal rank fusion and the Fisher linear discriminant, which were not provided. The prompt encourages complex multi-round logic chains and explicitly grades programs by structural depth, from single-pass manipulations through multi-stage conditional pipelines.

Programs are subject to six constraints: they must be task-universal with globally fixed constants, deterministic, free of any external model or learnable parameter, structurally distinct from every program already in the registry $\mathcal{R}$, and evaluated once on all $T$ tasks with no per-task sweep. Each proposal also records a one-sentence novelty claim and a hypothesis, which enter the history and let the proposer build on prior analyses rather than revisit ruled-out ones.

\subsection{Evaluator}
\label{sec:harness}

The evaluator is designed around the shortlist assumption. In production, a fast first-stage retriever such as BM25 or single-pass dense cosine narrows candidates to a shortlist of hundreds to low thousands, and TTC programs operate on this shortlist. Our evaluation corpora contain 46 to 438 documents per task, naturally representative of second-stage reranking, so the cost ratios measured here generalize directly to production shortlist sizes.

Programs are evaluated on $T{=}14$ retrieval tasks across four domains summarized in Table~\ref{tab:tasks}. The tasks span legal, financial, long-document, and general retrieval, with corpus sizes ranging from 46 to 438 documents and average document lengths from 73 to over 50{,}000 words. The search-phase encoder is \jinanano{} running natively on Apple Silicon via MLX.

\begin{table}[t]
\centering
\caption{The 14 Tier~1 retrieval tasks used for program evaluation during the agentic search, grouped by domain. $|Q|$: queries, $|D|$: documents, Avg.~$|d|$: mean document length in words.}
\label{tab:tasks}
\small
\setlength{\tabcolsep}{2pt}
\begin{tabular}{llrrr}
\toprule
\textbf{Task} & \textbf{Domain} & $|Q|$ & $|D|$ & \textbf{Avg.~$|d|$} \\
\midrule
AILACasedocs       & Legal     &     50 &   186 &  4{,}637 \\
AILAStatutes        & Legal     &     50 &    82 &    337 \\
BarExamQA           & Legal     &    117 &   116 &    109 \\
LegalSummarization  & Legal     &    284 &   438 &    102 \\
\midrule
FinanceBenchRetrieval & Financial &  150 &   145 &    230 \\
FinQARetrieval      & Financial & 1{,}138 &   380 &    660 \\
HC3FinanceRetrieval & Financial &    415 &   415 &    175 \\
\midrule
LEMBNarrativeQA     & Long-doc  & 10{,}449 & 355 & 50{,}474 \\
LEMBNeedle          & Long-doc  &     50 &   100 &    769 \\
LEMBPasskey          & Long-doc  &     50 &   100 &    759 \\
LEMBQMSum           & Long-doc  &  1{,}527 &  197 & 10{,}058 \\
LEMBSummScreenFD    & Long-doc  &    336 &   336 &  5{,}582 \\
LEMBWikimQA         & Long-doc  &    300 &   300 &  6{,}132 \\
\midrule
LIMITSmall          & General   &  1{,}000 &   46 &     73 \\
\bottomrule
\end{tabular}
\end{table}

Let $T_{\text{base}} = \sum_{t=1}^{T}(|Q_t| + |D_t|)$ denote the total number of texts encoded once across all $T{=}14$ tasks, and let $T_{\text{prog}}$ denote the additional encoder invocations a program makes at retrieval time. The \emph{cost ratio} $c = (T_{\text{base}} + T_{\text{prog}}) / T_{\text{base}}$ measures the multiplicative overhead in encoder forward passes relative to the single-pass baseline. Per query, the baseline $P_0$ runs in $\mathcal{O}(F + dN_c)$ and a program at cost ratio $c$ in $\mathcal{O}(cF + dN_c)$, where $F$ is the encoder forward-pass cost and $dN_c$ is the scoring multiply over $N_c$ shortlisted documents of dimension $d$. The $cF$ term dominates: $F$ is roughly 10 to 50\,ms for transformer encoders, whereas the $dN_c$ multiply takes about $0.05$\,ms at $N_c{=}1{,}000$ and $d{=}1{,}024$ on a single CPU thread.

\subsection{Memory}

The memory is structured rather than narrative, since free-text ledgers drift and accumulate contradictory lessons that the proposer cannot reliably weigh. The history $\mathcal{H}$ is a JSONL ledger with one record per program holding its generation index, per-task $\Delta$nDCG@10, win/tie/loss counts, cost ratio, parent, and the novelty and hypothesis strings, and each program's source carries a header with root-cause analysis of prior failures and task-level forensics against its parent, so the agent accumulates analytical depth by building on the analyses of its predecessors.

\subsection{Agentic loop}

The loop evaluates one program per generation on all $T$ tasks under a compute lock, with fixed constants and no parameter sweep, and appends the result to the history (Algorithm~\ref{alg:meta} in Appendix~\ref{app:trajectory}). A program enters the frontier if its mean $\Delta$nDCG@10 across the 14 discovery tasks is positive and exceeds every previously admitted program, and we record the mean, a win/tie/loss triplet at $\pm 0.001$, and the cost ratio. No held-out task or model is consulted during the search.

\subsection{Discovered programs}
\label{sec:discovery}
\label{sec:programs}

\begin{figure}[t]
\centering
\includegraphics[width=\columnwidth]{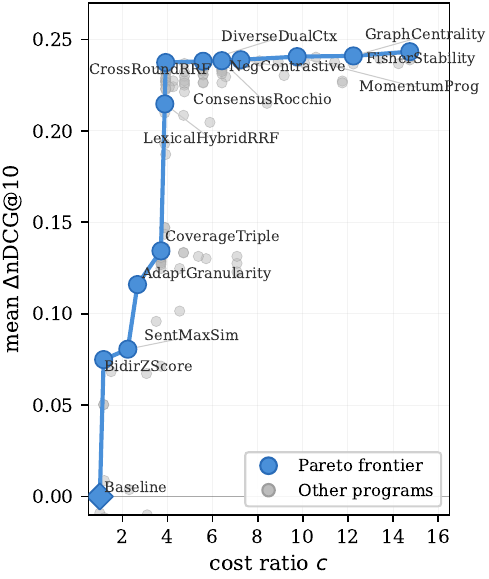}
\caption{Mean $\Delta$nDCG@10 on the 14 Tier~1 discovery tasks versus cost ratio $c$ for all 144 programs produced by the search. Blue circles mark the 12 Pareto-optimal programs; gray points are dominated. The curve interpolates the frontier from the baseline. These are in-search measurements; held-out validation appears in Section~\ref{sec:exp}.}
\label{fig:pareto_frontier}
\end{figure}

The search produced 144 programs across 144 generations. Figure~\ref{fig:pareto_frontier} shows the mean $\Delta$nDCG@10 against cost ratio $c$ for all 144 programs. Twelve lie on the Pareto frontier, spanning $c{=}1.2$ to $14.7$. Table~\ref{tab:frontier-unified} in Section~\ref{sec:exp} summarizes these twelve programs with both discovery-phase and held-out W/T/L counts. The full algorithmic details and connections to the retrieval literature appear in Appendix~\ref{app:programs}.

Four structural families emerge along the frontier. At $c \le 2.7$ programs exploit sub-document decomposition and cross-perspective scoring; at $c \approx 3.7$ to $3.9$ lexical channels and reciprocal rank fusion appear, converging to hybrid sparse-dense retrieval and Rocchio feedback \citep{rocchio1971relevance,cormack2009reciprocal}; at $c \approx 5$ to $7$ contextual query expansion and consensus filtering add nonlinear query-document interactions; and at $c > 9$ multi-round expansion and algebraic channels, including a rediscovery of the Fisher linear discriminant \citep{fisher1936lda}, add marginal gains. The steepest gains fall between $c{=}1$ and $c{=}4$ as perspective swap, granularity, lexical matching, and iterative refinement are introduced, after which further rounds refine an already-converged query with diminishing signal.

Appendix~\ref{app:trajectory} visualizes the search trajectory through program space across the 144 generations, showing the move from cheap geometric programs toward the more expensive multi-round programs while the frontier stays among the higher-scoring neighbourhoods.

\section{Experimental Results}
\label{sec:exp}

\subsection{Setup}
\label{sec:exp-setup}

The twelve frontier programs, discovered on \jinanano{} over the 14 Tier~1 tasks with corpora of 46 to 438 documents, are run unmodified alongside the cosine baseline $P_0$ on held-out models and tasks. The small discovery corpora keep per-generation cost low enough for 144 generations, and transfer to larger corpora is validated below.

The four evaluation encoders span three architecturally distinct families: \embgemma{} \citep{vera2025embeddinggemma}, a 303\,M decoder built on Gemma~3 with no adapters; \qwenthreeemb{}, a 600\,M Qwen3 decoder, also without adapters; and the Jina pair \jinanano{}, the 239\,M discovery model, and \jinasmall{}, a 568\,M variant with multi-LoRA adapters. \embgemma{} and \qwenthreeemb{} share no training data, tokenizer, or adapter design with the discovery model. The nineteen held-out tasks are drawn from MMTEB Tier~2 and Tier~3, none in the discovery set, with corpora up to roughly 100\,K documents spanning summarization, legal and medical QA, fact verification, argument retrieval, and temporal and commonsense reasoning. All models run via MLX in float16 under the MTEB~v2.8.4 protocol, baseline reproduction matches published nDCG@10 within 0.4 points (Appendix~\ref{app:baseline}), and the full matrix of $4 \times 19 \times 13 = 988$ cells has complete coverage.

\begin{table*}[t]
\centering
\caption{The twelve Pareto-optimal frontier programs with win/tie/loss counts on the 14 discovery tasks and 19 held-out evaluation tasks across four encoder families. Discovery W/T/L reflects in-search performance on \jinanano{}; the four right columns validate generalization to unseen tasks and models. Algorithmic details for each program appear in Appendix~\ref{app:programs}. Threshold $\pm 0.001$.}
\label{tab:frontier-unified}
\small
\setlength{\tabcolsep}{4pt}
\begin{tabular}{@{}lccccccc@{}}
\toprule
& & \multicolumn{1}{c}{Discovery (14 tasks)} & \multicolumn{4}{c}{Held-out evaluation (19 tasks)} \\
\cmidrule(lr){3-3}\cmidrule(lr){4-7}
\textbf{Program} & $c$ & \jinanano{} & \jinanano{} & \jinasmall{} & \embgemma{} & \qwenthreeemb{} \\
\midrule
\textsc{BidirZScore}     & $1.2$ & $\mathbf{13/1/0}$ & $6/2/11$ & $5/2/12$  & $10/5/4$  & $9/2/8$ \\
\textsc{SentMaxSim}      & $2.2$ & $11/1/2$ & $6/1/12$ & $4/1/14$  & $9/4/6$   & $10/1/8$ \\
\textsc{AdaptGran}       & $2.7$ & $\mathbf{13/1/0}$ & $7/1/11$ & $5/1/13$  & $\mathbf{12/3/4}$  & $11/1/7$ \\
\textsc{CovTriple}       & $3.7$ & $\mathbf{13/1/0}$ & $6/3/10$ & $5/3/11$  & $11/3/5$  & $\mathbf{13/1/5}$ \\
\textsc{LexHybridRRF}    & $3.9$ & $\mathbf{13/1/0}$ & $9/2/8$  & $8/1/10$  & $11/3/5$  & $\mathbf{13/1/5}$ \\
\textsc{CrossRoundRRF}   & $3.9$ & $12/0/2$ & $\mathbf{10/1/8}$ & $\mathbf{9/0/10}$  & $\mathbf{12/3/4}$  & $10/3/6$ \\
\textsc{DiverseDualCtx}  & $5.6$ & $12/1/1$ & $9/2/8$  & $\mathbf{9/0/10}$  & $\mathbf{12/3/4}$  & $11/1/7$ \\
\textsc{ConsRocchio}     & $6.4$ & $12/1/1$ & $9/2/8$  & $\mathbf{9/0/10}$  & $12/2/5$  & $12/0/7$ \\
\textsc{NegContrast}     & $7.2$ & $12/1/1$ & $9/2/8$  & $\mathbf{9/0/10}$  & $12/2/5$  & $12/1/6$ \\
\textsc{MomentumProg}    & $9.8$ & $12/1/1$ & $9/2/8$  & $\mathbf{9/0/10}$  & $12/2/5$  & $12/1/6$ \\
\textsc{GraphCent}       & $12.2$ & $12/1/1$ & $\mathbf{10/1/8}$ & $\mathbf{9/0/10}$ & $12/2/5$  & $12/1/6$ \\
\textsc{FisherStab}      & $14.7$ & $12/1/1$ & $9/2/8$  & $\mathbf{9/0/10}$ & $12/2/5$  & $12/1/6$ \\
\bottomrule
\end{tabular}
\end{table*}

\begin{figure*}[t]
\centering
\includegraphics[width=\textwidth]{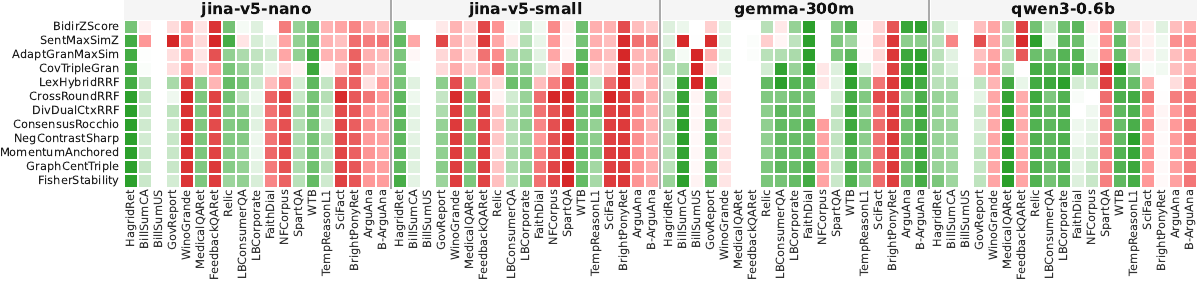}
\caption{Cross-model and cross-task validation of all twelve Pareto-optimal frontier programs. Each column within a model panel corresponds to one of nineteen evaluation tasks; each row corresponds to one of twelve frontier programs. Cell color encodes $\Delta$nDCG@10 relative to the per-model cosine baseline: green is improvement, red is regression. Color is normalized per task (column) to maximize contrast. Programs were discovered on \jinanano{} and transferred without modification to \jinasmall{}, \embgemma{}, and \qwenthreeemb{}. Of 76 model--task pairs, 52 admit at least one frontier program that improves over the baseline.}
\label{fig:xmodel-heatmap}
\end{figure*}

\subsection{Cross-model transfer}
\label{sec:xmodel-model}

Table~\ref{tab:frontier-unified} reports W/T/L counts for each frontier program on both the 14 discovery tasks and 19 held-out evaluation tasks across all four models. Figure~\ref{fig:xmodel-heatmap} provides the per-cell $\Delta$nDCG@10.\footnote{$\Delta$nDCG@10 for a program on a single task is the per-query nDCG@10 of that program minus the per-query nDCG@10 of the cosine baseline $P_0$, averaged over the task's queries, so that one (model,~task) cell yields one $\Delta$nDCG@10 value. Unless stated otherwise, the held-out median and win-rate for a program pool these per-cell values over all (model,~task) cells, with the median taken over the pooled values and the win-rate defined as the fraction of cells with $\Delta$nDCG@10 $> 0.001$.} The central question is whether any single discovered program, applied without task-specific selection, improves retrieval across unseen encoder families and tasks.

\textsc{LexHybridRRF}, the lowest-cost program that combines lexical and semantic channels, is our representative choice from discovery-phase properties alone. Applied without modification it wins a $54\%$ rate over the 76 held-out cells with a positive pooled median, nine of nineteen tasks are unanimously positive across all four encoders, and none is unanimously negative, so the programs generalize beyond the discovery encoder without oracle selection. Transfer is in fact strongest on the encoders never seen during discovery: an oracle over the frontier lifts 52 of 76 cells, led by \embgemma{} on 15 of 19 tasks at a mean $+0.031$ and \qwenthreeemb{} on 15 of 19 at $+0.017$.

This foreign-family advantage is mechanistic. When an encoder lacks the LoRA adapters that five programs request, the operations that survive are geometric, namely z-scoring, sub-document decomposition, and centroid feedback, and these depend on the cosine structure of the embedding space rather than on model-specific training artifacts, so they carry to representation spaces never seen during discovery.

\subsection{Cross-task transfer}
\label{sec:xmodel-task}

The winning programs map structural mechanisms to task families. Sub-document granularity gives the strongest lifts on long documents: \textsc{CoverageTriple} scores by multi-granularity MaxSim over sentence embeddings and surfaces local evidence that a single document vector smooths over, gaining $+0.116$ to $+0.143$ on \textsc{BIRCO-WTB} and driving the lifts on the \textsc{BillSum} and \textsc{GovReport} legislative tasks. Lexical-hybrid fusion helps terminology-heavy domains, where \textsc{LexHybridRRF} recovers the exact-match signal of statutory phrases that dense embeddings underweight, gaining $+0.039$ to $+0.067$ on the legal contract tasks.

Where transfer is model-dependent it tracks encoder geometry rather than the program. Counter-argument retrieval on \textsc{ArguAna} gains $+0.081$ to $+0.120$ on \embgemma{} alone, since separating supporting from contradicting passages depends on the backbone pre-training objective, and \textsc{MedicalQARetrieval} gains $+0.088$ to $+0.156$ on the three encoders whose space encodes medical similarity. The boundary is the encoder, not the recipe: test-time compute amplifies the latent signal a space already carries but does not create it, so near-saturated tasks leave no headroom while a sufficiently receptive encoder such as \qwenthreeemb{} extends the benefit even to short-document tasks that are flat elsewhere.

\subsection{Baselines, robustness, and deployment cost}
\label{sec:baselines-main}

A matched-budget learned head is the natural training-time alternative, and it does not transfer. Trained on the 14 discovery tasks as a linear, low-rank, or MLP map over the frozen embeddings, the best head improves in-domain retrieval by $+0.20$ to $+0.25$ nDCG@10 yet falls below baseline on every held-out encoder, since a learned metric memorizes the discovery domains. Test-time compute, not a small learned metric, is therefore the axis that generalizes with no per-domain labels (Appendix~\ref{app:headbaseline}).

The gains come from the structural operations rather than from hybrid fusion alone. Against classical feedback, \textsc{SoftCentroid}, a minimal centroid-replacement program that isolates the PRF mechanism, exceeds the best of sixteen Rocchio and VPRF configurations on every encoder by $+2.31$ to $+6.00$ nDCG@10 at $p<10^{-4}$ (Appendix~\ref{app:rocchio-comparison}). A vanilla baseline that fuses BM25 with dense cosine through the same reciprocal rank fusion, stripped of z-scoring, granularity, and feedback, reaches only a $29\%$ held-out win-rate, whereas \textsc{LexHybridRRF} on the same two channels reaches $55\%$, so the fusion of lexical and dense scores alone does not explain the transfer.

A paired bootstrap with $10{,}000$ resamples confirms the held-out gains are majority-positive, with \textsc{LexHybridRRF} recording $5$ to $7$ significant wins against $7$ significant losses per encoder and \qwenthreeemb{} and \embgemma{} the most favourable. As with classical pseudo-relevance feedback, a few queries drift, so we report median and win-rate alongside the mean, and the discovered \textsc{FisherStability} rank-stability channel and \textsc{ConsensusRocchio} consensus filtering temper this drift directly.

Deployment cost depends on whether the re-encoding is amortizable. The cost ratio $c$ sums query- and document-side encoder calls, but a micro-benchmark over the full versus half query set separates them. \textsc{BidirZScore} and the granularity programs issue no per-query encoder call and re-encode the corpus once at index time, so their query-time cost ratio is close to $1$, whereas the query-expansion family re-encodes per query.
\label{sec:cost-honest}
The algebraic operations are negligible, $0.011$\,s against the $40$\,s of encoding on the measured cell, so the index-amortizable programs at the cheap end of the frontier are the most practical to deploy.

\section{Conclusion}
\label{sec:conclusion}

Small, LLM-distilled embedding models inherit usable test-time-compute potential, and a frozen encoder converts inference compute into retrieval gains that transfer to unseen encoder families with no per-domain labels. An agentic search over 144 generations discovers twelve Pareto-optimal programs that require no auxiliary models, no parameters trained at deployment, and no per-task tuning, selected on 14 discovery tasks and validated on 19 held-out tasks across four encoder families (Table~\ref{tab:frontier-unified}). On held-out transfer a single fixed program helps the typical task, with a positive median $\Delta$nDCG@10 and a $54$ to $57\%$ win-rate at $c{\ge}4$, and the gains are largest on the encoder families never seen during discovery. Structural inference-time compute is the only one of the two compute axes that transfers this way, since a matched-budget projection head trained on the same discovery tasks improves in-domain retrieval by $+0.20$ to $+0.25$ nDCG@10 yet falls below baseline on every held-out encoder, a learned metric having memorized the discovery domains. The discovered programs exploit geometric properties of LLM-descended embedding spaces, using sub-document granularity for long documents, lexical-hybrid fusion for terminology-heavy domains, and centroid feedback for general retrieval, and the index-amortizable variants at the cheap end of the frontier are the most practical to deploy. Test-time compute thus opens a scaling axis orthogonal to model size, one that any frozen embedder can use, with the steepest gains below $4{\times}$ and the cheapest programs reducible to a one-time index pass.

\section*{Limitations}

The gains concentrate on the unseen \embgemma{} and \qwenthreeemb{} families and are more modest on the discovery family, which indicates that the programs exploit general embedding geometry rather than discovery-model artifacts. We report transfer by median and win-rate, since a few query-drift cases keep the arithmetic mean near the baseline. The learned-head baseline is deliberately small and trained only on the 14 discovery tasks, so the comparison is confined to the matched-budget, no-in-domain-label regime, and the program search itself consumes labels on those tasks, so the method is label-free at deployment rather than end-to-end. Significance is reported for three representative programs spanning the cost range, full-corpus scaling for four BEIR tasks, and a multilingual evaluation of the geometric core is left to future work, since the lexical-hybrid programs carry Latin-script assumptions in their tokenizer and sentence splitter. Encoders at 4B scale and multimodal encoders are untested, and the most expensive program needs $14.7{\times}$ baseline encoder calls, though the steepest gains concentrate below $4{\times}$.

\bibliography{refs}

\begin{thebibliography}{38}
\providecommand{\natexlab}[1]{#1}

\bibitem[{Akram et~al.(2026)Akram, Sturua, Havriushenko, Herreros, G{\"u}nther,
  Werk, and Xiao}]{akram2026jinav5}
Mohammad~Kalim Akram, Saba Sturua, Nastia Havriushenko, Quentin Herreros,
  Michael G{\"u}nther, Maximilian Werk, and Han Xiao. 2026.
\newblock jina-embeddings-v5-text: Task-targeted embedding distillation.
\newblock \emph{arXiv preprint arXiv:2602.15547}.

\bibitem[{Brown et~al.(2024)Brown, Juravsky, Ehrlich, Clark, Le, R{\'e}, and
  Mirhoseini}]{brown2024monkeys}
Bradley Brown, Jordan Juravsky, Ryan Ehrlich, Ronald Clark, Quoc~V. Le,
  Christopher R{\'e}, and Azalia Mirhoseini. 2024.
\newblock Large language monkeys: Scaling inference compute with repeated
  sampling.
\newblock \emph{arXiv preprint arXiv:2407.21787}.

\bibitem[{Cormack et~al.(2009)Cormack, Clarke, and
  Buettcher}]{cormack2009reciprocal}
Gordon~V. Cormack, Charles L.~A. Clarke, and Stefan Buettcher. 2009.
\newblock \href {https://doi.org/10.1145/1571941.1572114} {Reciprocal rank
  fusion outperforms condorcet and individual rank learning methods}.
\newblock In \emph{SIGIR}, pages 758--759.

\bibitem[{Fisher(1936)}]{fisher1936lda}
Ronald~A. Fisher. 1936.
\newblock \href {https://doi.org/10.1111/j.1469-1809.1936.tb02137.x} {The use
  of multiple measurements in taxonomic problems}.
\newblock \emph{Annals of Eugenics}, 7(2):179--188.

\bibitem[{Formal et~al.(2021)Formal, Piwowarski, and
  Clinchant}]{formal2021splade}
Thibault Formal, Benjamin Piwowarski, and St{\'e}phane Clinchant. 2021.
\newblock \href {https://doi.org/10.1145/3404835.3463098} {{SPLADE}: Sparse
  lexical and expansion model for first stage ranking}.
\newblock In \emph{SIGIR}, pages 2288--2292.

\bibitem[{Gao et~al.(2023)Gao, Ma, Lin, and Callan}]{gao2023hyde}
Luyu Gao, Xueguang Ma, Jimmy Lin, and Jamie Callan. 2023.
\newblock \href {https://doi.org/10.18653/v1/2023.acl-long.99} {Precise
  zero-shot dense retrieval without relevance labels}.
\newblock In \emph{ACL}, pages 1762--1777.

\bibitem[{Khattab and Zaharia(2020)}]{khattab2020colbert}
Omar Khattab and Matei Zaharia. 2020.
\newblock \href {https://doi.org/10.1145/3397271.3401075} {{ColBERT}: Efficient
  and effective passage search via contextualized late interaction over
  {BERT}}.
\newblock In \emph{SIGIR}, pages 39--48.

\bibitem[{Lavrenko and Croft(2001)}]{lavrenko2001rm3}
Victor Lavrenko and W.~Bruce Croft. 2001.
\newblock \href {https://doi.org/10.1145/383952.383972} {Relevance-based
  language models}.
\newblock In \emph{SIGIR}, pages 120--127.

\bibitem[{Lee et~al.(2025)Lee, Roy, Xu, Raiman, Shoeybi, Catanzaro, and
  Ho}]{lee2024nvembed}
Chankyu Lee, Rajarshi Roy, Mengjie Xu, Jonathan Raiman, Mohammad Shoeybi, Bryan
  Catanzaro, and Wei Ho. 2025.
\newblock \href
  {https://openreview.net/forum?id=c4bf73386022473a652a18941e9ea6f8}
  {{NV-Embed}: Improved techniques for training {LLMs} as generalist embedding
  models}.
\newblock In \emph{ICLR}.
\newblock ArXiv:2405.17428.

\bibitem[{Lehman et~al.(2024)Lehman, Gordon, Jain, Ndousse, Yeh, and
  Stanley}]{lehman2023elm}
Joel Lehman, Jonathan Gordon, Shawn Jain, Kamal Ndousse, Cathy Yeh, and
  Kenneth~O. Stanley. 2024.
\newblock \href {https://doi.org/10.1007/978-981-99-3814-8_11} {Evolution
  through large models}.
\newblock In Wolfgang Banzhaf, Penousal Machado, and Mengjie Zhang, editors,
  \emph{Handbook of Evolutionary Machine Learning}, Genetic and Evolutionary
  Computation. Springer.

\bibitem[{Li et~al.(2024)Li, Qin, Xiao, Chen, Luo, Shao, Lian, and
  Liu}]{li2024bgeenicl}
Chaofan Li, Minghao Qin, Shitao Xiao, Jianlyu Chen, Kun Luo, Yingxia Shao, Defu
  Lian, and Zheng Liu. 2024.
\newblock Making text embedders few-shot learners.
\newblock \emph{arXiv preprint arXiv:2409.15700}.

\bibitem[{Li et~al.(2023)Li, Mourad, Zhuang, Koopman, and Zuccon}]{li2023vprf}
Hang Li, Ahmed Mourad, Shengyao Zhuang, Bevan Koopman, and Guido Zuccon. 2023.
\newblock \href {https://doi.org/10.1145/3570724} {Pseudo relevance feedback
  with deep language models and dense retrievers: Successes and pitfalls}.
\newblock \emph{ACM Transactions on Information Systems}, 41(3):1--40.

\bibitem[{Li et~al.(2025{\natexlab{a}})Li, Wang, Koopman, and
  Zuccon}]{li2025promptprf}
Hang Li, Xiao Wang, Bevan Koopman, and Guido Zuccon. 2025{\natexlab{a}}.
\newblock Pseudo relevance feedback is enough to close the gap between small
  and large dense retrieval models.
\newblock \emph{arXiv preprint arXiv:2503.14887}.

\bibitem[{Li et~al.(2025{\natexlab{b}})Li, Zhuang, Koopman, and
  Zuccon}]{li2025llmvprf}
Hang Li, Shengyao Zhuang, Bevan Koopman, and Guido Zuccon. 2025{\natexlab{b}}.
\newblock {LLM-VPRF}: Large language model based vector pseudo relevance
  feedback.
\newblock \emph{arXiv preprint arXiv:2504.01448}.

\bibitem[{Li et~al.(2022)Li, Zhuang, Mourad, Ma, Lin, and
  Zuccon}]{li2022anceprfrepro}
Hang Li, Shengyao Zhuang, Ahmed Mourad, Xueguang Ma, Jimmy Lin, and Guido
  Zuccon. 2022.
\newblock \href {https://doi.org/10.1007/978-3-030-99736-6_40} {Improving query
  representations for dense retrieval with pseudo relevance feedback: A
  reproducibility study}.
\newblock In \emph{ECIR}, pages 599--612.
\newblock ArXiv:2112.06400.

\bibitem[{Ma et~al.(2024)Ma, Wang, Yang, Wei, and Lin}]{ma2024repllama}
Xueguang Ma, Liang Wang, Nan Yang, Furu Wei, and Jimmy Lin. 2024.
\newblock Fine-tuning {LLaMA} for multi-stage text retrieval.
\newblock In \emph{SIGIR}.
\newblock ArXiv:2310.08319.

\bibitem[{Mackie et~al.(2023)Mackie, Chatterjee, and Dalton}]{mackie2023grfprf}
Iain Mackie, Shubham Chatterjee, and Jeffrey Dalton. 2023.
\newblock Generative and pseudo-relevant feedback for sparse, dense and learned
  sparse retrieval.
\newblock \emph{arXiv preprint arXiv:2305.07477}.

\bibitem[{Meng et~al.(2024)Meng, Liu, Yavuz, Agarwal, Tu, Yu, Zhang, Chen, and
  Raghavan}]{meng2024sfr}
Rui Meng, Ye~Liu, Semih Yavuz, Rishabh Agarwal, Lifu Tu, Ning Yu, Jiacheng
  Zhang, Zhengdong Chen, and Hetal Raghavan. 2024.
\newblock {SFR-Embedding-2}: Advanced text embeddings with multi-stage
  training.
\newblock Salesforce AI Research.
\newblock \url{https://huggingface.co/Salesforce/SFR-Embedding-2_R}.

\bibitem[{Muennighoff et~al.(2025)Muennighoff, Su, Wang, Yang, Wei, and
  Shi}]{muennighoff2024gritlm}
Niklas Muennighoff, Hongjin Su, Liang Wang, Nan Yang, Furu Wei, and Tao Shi.
  2025.
\newblock \href {https://openreview.net/forum?id=BC4lIvfSzv} {Generative
  representational instruction tuning}.
\newblock In \emph{ICLR}.
\newblock ArXiv:2402.09906.

\bibitem[{Naseri et~al.(2021)Naseri, Dalton, Yates, and Allan}]{naseri2021ceqe}
Shahrzad Naseri, Jeffrey Dalton, Andrew Yates, and James Allan. 2021.
\newblock \href {https://doi.org/10.1007/978-3-030-72113-8_31} {{CEQE}:
  Contextualized embeddings for query expansion}.
\newblock In \emph{ECIR}, pages 467--482.

\bibitem[{Novikov et~al.(2025)Novikov, V{\~u}, Eisenberger, Dupont, Huang,
  Wagner, Shirobokov, Kozlovskii, Ruiz, Mehrabian, Kumar, See, Chaudhuri,
  Holland, Davies, Nowozin, Kohli, and Balog}]{novikov2025alphaevolve}
Alexander Novikov, Ng{\^a}n V{\~u}, Marvin Eisenberger, Emilien Dupont, Po-Sen
  Huang, Adam~Zsolt Wagner, Sergey Shirobokov, Borislav Kozlovskii, Francisco
  J.~R. Ruiz, Abbas Mehrabian, M.~Pawan Kumar, Abigail See, Swarat Chaudhuri,
  George Holland, Alex Davies, Sebastian Nowozin, Pushmeet Kohli, and Matej
  Balog. 2025.
\newblock {AlphaEvolve}: A coding agent for scientific and algorithmic
  discovery.
\newblock \emph{arXiv preprint arXiv:2506.13131}.

\bibitem[{Robertson et~al.(1995)Robertson, Walker, Jones, Hancock-Beaulieu, and
  Gatford}]{robertson1995okapi}
Stephen~E. Robertson, Steve Walker, Susan Jones, Micheline~M. Hancock-Beaulieu,
  and Mike Gatford. 1995.
\newblock {Okapi at TREC-3}.
\newblock In \emph{Proceedings of the Third Text REtrieval Conference (TREC-3),
  NIST Special Publication 500-225}, pages 109--126. National Institute of
  Standards and Technology.

\bibitem[{Rocchio(1971)}]{rocchio1971relevance}
Joseph~J. Rocchio. 1971.
\newblock Relevance feedback in information retrieval.
\newblock In Gerard Salton, editor, \emph{The {SMART} Retrieval System:
  Experiments in Automatic Document Processing}, pages 313--323. Prentice-Hall,
  Englewood Cliffs, NJ.

\bibitem[{Romera-Paredes et~al.(2024)Romera-Paredes, Barekatain, Novikov,
  Balog, Kumar, Dupont, Ruiz, Ellenberg, Wang, Fawzi, Kohli, and
  Fawzi}]{romera2024funsearch}
Bernardino Romera-Paredes, Mohammadamin Barekatain, Alexander Novikov, Matej
  Balog, M.~Pawan Kumar, Emilien Dupont, Francisco J.~R. Ruiz, Jordan~S.
  Ellenberg, Pengming Wang, Omar Fawzi, Pushmeet Kohli, and Alhussein Fawzi.
  2024.
\newblock \href {https://doi.org/10.1038/s41586-023-06924-6} {Mathematical
  discoveries from program search with large language models}.
\newblock \emph{Nature}, 625(7995):468--475.

\bibitem[{Santhanam et~al.(2022)Santhanam, Khattab, Saad-Falcon, Potts, and
  Zaharia}]{santhanam2022colbertv2}
Keshav Santhanam, Omar Khattab, Jon Saad-Falcon, Christopher Potts, and Matei
  Zaharia. 2022.
\newblock \href {https://doi.org/10.18653/v1/2022.naacl-main.272} {{ColBERTv2}:
  Effective and efficient retrieval via lightweight late interaction}.
\newblock In \emph{NAACL}, pages 3715--3734.

\bibitem[{Snell et~al.(2025)Snell, Lee, Xu, and Kumar}]{snell2024scalingttc}
Charlie Snell, Jaehoon Lee, Kelvin Xu, and Aviral Kumar. 2025.
\newblock \href {https://openreview.net/forum?id=4FWAwZtd2n} {Scaling {LLM}
  test-time compute optimally can be more effective than scaling parameters for
  reasoning}.
\newblock In \emph{ICLR}.
\newblock ArXiv:2408.03314.

\bibitem[{Tu et~al.(2025)Tu, Su, Zhou, Liu, Lin, Liu, and Ai}]{tu2025gprf}
Yiteng Tu, Weihang Su, Yujia Zhou, Yiqun Liu, Fen Lin, Qin Liu, and Qingyao Ai.
  2025.
\newblock Generalized pseudo-relevance feedback.
\newblock \emph{arXiv preprint arXiv:2510.25488}.

\bibitem[{Uzan et~al.(2026)Uzan, Yehudai, Pony, Shnarch, and
  Gera}]{uzan2025gqr}
Omri Uzan, Asaf Yehudai, Roi Pony, Eyal Shnarch, and Ariel Gera. 2026.
\newblock \href {https://openreview.net/forum?id=4GRsedu43K} {Guided query
  refinement: Multimodal hybrid retrieval with test-time optimization}.
\newblock In \emph{ICLR}.
\newblock ArXiv:2510.05038.

\bibitem[{Vera et~al.(2025)Vera, Dua, Zhang, Salz, Mullins
  et~al.}]{vera2025embeddinggemma}
Henrique~Schechter Vera, Sahil Dua, Biao Zhang, Daniel Salz, Ryan Mullins, and
  1 others. 2025.
\newblock {EmbeddingGemma}: Powerful and lightweight text representations.
\newblock \emph{arXiv preprint arXiv:2509.20354}.

\bibitem[{Wang et~al.(2024)Wang, Yang, Huang, Yang, Majumder, and
  Wei}]{wang2024e5mistral}
Liang Wang, Nan Yang, Xiaolong Huang, Linjun Yang, Rangan Majumder, and Furu
  Wei. 2024.
\newblock Improving text embeddings with large language models.
\newblock In \emph{ACL}.
\newblock ArXiv:2401.00368.

\bibitem[{Wang et~al.(2023{\natexlab{a}})Wang, Yang, and
  Wei}]{wang2023query2doc}
Liang Wang, Nan Yang, and Furu Wei. 2023{\natexlab{a}}.
\newblock \href {https://doi.org/10.18653/v1/2023.emnlp-main.585} {{Query2doc}:
  Query expansion with large language models}.
\newblock In \emph{EMNLP}, pages 9414--9423.

\bibitem[{Wang et~al.(2021)Wang, Macdonald, Tonellotto, and
  Ounis}]{wang2021colbertprf}
Xiao Wang, Craig Macdonald, Nicola Tonellotto, and Iadh Ounis. 2021.
\newblock \href {https://doi.org/10.1145/3471158.3472250} {Pseudo-relevance
  feedback for multiple representation dense retrieval}.
\newblock In \emph{Proceedings of the 2021 ACM SIGIR International Conference
  on Theory of Information Retrieval}, pages 297--306.

\bibitem[{Wang et~al.(2023{\natexlab{b}})Wang, Wei, Schuurmans, Le, Chi,
  Narang, Chowdhery, and Zhou}]{wang2023selfconsistency}
Xuezhi Wang, Jason Wei, Dale Schuurmans, Quoc~V. Le, Ed~H. Chi, Sharan Narang,
  Aakanksha Chowdhery, and Denny Zhou. 2023{\natexlab{b}}.
\newblock \href {https://openreview.net/forum?id=1PL1NIMMrw} {Self-consistency
  improves chain of thought reasoning in language models}.
\newblock In \emph{ICLR}.

\bibitem[{Wu et~al.(2025)Wu, Sun, Li, Welleck, and
  Yang}]{wu2024inferencescaling}
Yangzhen Wu, Zhiqing Sun, Shanda Li, Sean Welleck, and Yiming Yang. 2025.
\newblock Inference scaling laws: An empirical analysis of compute-optimal
  inference for problem-solving with language models.
\newblock In \emph{ICLR}.
\newblock ArXiv:2408.00724.

\bibitem[{Xiao et~al.(2026)Xiao, Ma, Gu, Chen, Chen, Ordonez, and
  Mohan}]{xiao2025metaembed}
Zilin Xiao, Qi~Ma, Mengting Gu, Chun-cheng~Jason Chen, Xintao Chen, Vicente
  Ordonez, and Vijai Mohan. 2026.
\newblock \href {https://openreview.net/forum?id=yKDqg9HwZX} {{MetaEmbed}:
  Scaling multimodal retrieval at test-time with flexible late interaction}.
\newblock In \emph{ICLR}.
\newblock ArXiv:2509.18095; Oral.

\bibitem[{Yu et~al.(2021)Yu, Xiong, and Callan}]{yu2021anceprf}
HongChien Yu, Chenyan Xiong, and Jamie Callan. 2021.
\newblock \href {https://doi.org/10.1145/3459637.3482124} {Improving query
  representations for dense retrieval with pseudo relevance feedback}.
\newblock In \emph{CIKM}, pages 3592--3596.

\bibitem[{Zhang et~al.(2025)Zhang, Li, Long, Zhang, Lin, Yang, Xie, Yang, Liu,
  Lin, Huang, and Zhou}]{zhang2025qwen3embedding}
Yanzhao Zhang, Mingxin Li, Dingkun Long, Xin Zhang, Huan Lin, Baosong Yang,
  Pengjun Xie, An~Yang, Dayiheng Liu, Junyang Lin, Fei Huang, and Jingren Zhou.
  2025.
\newblock {Qwen3 Embedding}: Advancing text embedding and reranking through
  foundation models.
\newblock \emph{arXiv preprint arXiv:2506.05176}.

\bibitem[{Zhuang et~al.(2024)Zhuang, Ma, Koopman, Lin, and
  Zuccon}]{zhuang2024promptreps}
Shengyao Zhuang, Xueguang Ma, Bevan Koopman, Jimmy Lin, and Guido Zuccon. 2024.
\newblock \href {https://doi.org/10.18653/v1/2024.emnlp-main.250}
  {{PromptReps}: Prompting large language models to generate dense and sparse
  representations for zero-shot document retrieval}.
\newblock In \emph{EMNLP}, pages 4375--4391.

\end{thebibliography}

\appendix
\nolinenumbers

\section{Program Search Trajectory}
\label{app:trajectory}

\begin{figure*}[t]
\centering
\includegraphics[width=\textwidth]{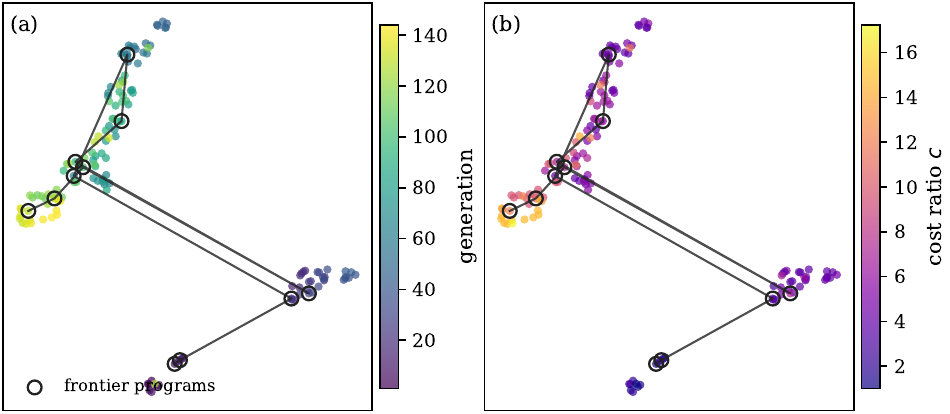}
\caption{Two-dimensional embedding of all 144 searched programs, positioned by their vectors of per-task $\Delta$nDCG@10 on the discovery tasks, so that programs which help the same tasks lie close together. The left panel colors points by generation and the right panel by cost ratio $c$. Circled points are the Pareto-optimal frontier programs, connected in ascending cost to trace the search's evolution path. The search migrates from cheap geometric programs to expensive multi-round programs while remaining among the higher-scoring regions.}
\label{fig:program_umap}
\end{figure*}

Figure~\ref{fig:program_umap} embeds all 144 searched programs in two dimensions from their vectors of per-task $\Delta$nDCG@10 on the discovery tasks, so that programs which help the same tasks lie close together. The search opens with cheap geometric programs in one region and migrates toward the more expensive multi-round programs as generations proceed, and the frontier programs trace a path that climbs in cost while staying among the higher-scoring neighbourhoods.

\begin{algorithm}[t]
\caption{Agentic embedding-program generation loop.}
\label{alg:meta}
\begin{algorithmic}[1]
\State $\mathcal{R} \gets \{P_0\}$, $\mathcal{F} \gets \{P_0\}$, $\mathcal{H} \gets \emptyset$
\For{$g = 1, 2, \dots, G$}
  \State $P^{\text{new}} \gets \textsc{Propose}(\mathcal{F}, \mathcal{H})$
  \State $\mathcal{R} \gets \mathcal{R} \cup \{P^{\text{new}}\}$
  \For{$t \in \{t_1, \dots, t_T\}$}
    \State $\Delta_t \gets \textsc{Evaluate}(P^{\text{new}}, t)$
  \EndFor
  \State $\mathcal{F} \gets \textsc{FrontierUpdate}(\mathcal{F}, P^{\text{new}}, \{\Delta_t\})$
  \State $\mathcal{H} \gets \mathcal{H} \cup \{(P^{\text{new}}, \{\Delta_t\}_t, c)\}$
\EndFor
\State \Return $\mathcal{F}$, $\mathcal{R}$, $\mathcal{H}$
\end{algorithmic}
\end{algorithm}

\section{Frontier Program Descriptions}
\label{app:programs}

This appendix describes each of the twelve Pareto-optimal frontier programs in order of increasing cost ratio $c$. Full Python source code for all programs is provided in the supplementary material. The geometric core shared by the programs, namely z-scoring, centroid feedback, and similarity over re-encoded chunks, inspects no language-specific feature and is language-agnostic, with \textsc{BidirZScore} the purely geometric example. The lexical-hybrid and sentence-granularity programs are less portable, since their tokenizer matches ASCII word characters and their sentence splitter assumes Latin terminal punctuation, so their lexical channels degrade on non-Latin scripts while the geometric channels continue to apply.

\subsection{\textsc{BidirZScore} ($c{=}1.2$)}
\label{app:bidir}

The cheapest frontier program. It re-encodes all $N$ documents with the query-side LoRA adapter \texttt{retrieval.query}, producing a second similarity matrix $\mathbf{S}'$ in which each document is scored as if it were a query against the original query embeddings. The original and reversed matrices are summed after per-column z-score normalization, converting raw cosine similarities into standardized deviates that measure how many standard deviations above its per-document mean a given query scores.

\subsection{\textsc{SentMaxSim} ($c{=}2.2$)}
\label{app:sentmaxsim}

Splits each document into sentences, encodes all sentences in a single batch, and scores each document by the maximum similarity between the query and any of its constituent sentences. This reproduces ColBERT's MaxSim operator \citep{khattab2020colbert,santhanam2022colbertv2} at sentence granularity (MaxSim was a seeded inspiration, Section~\ref{sec:proposer}); the original applies the same aggregation at the token level for late interaction. ColBERT computes MaxSim over per-token embeddings and requires multi-vector storage; SentMaxSim operates over sentence-level single-vector embeddings within the standard single-vector retrieval framework.

\subsection{\textsc{AdaptGranularity} ($c{=}2.7$)}
\label{app:adaptgran}

Decomposes documents at two levels, paragraphs and sentences, encoding both granularities in separate batches. Each granularity produces a MaxSim channel. An adaptive selector takes the element-wise maximum across channels after z-score normalization. The dual-granularity design recovers evidence in long-document tasks where paragraph-level matching captures context that sentence-level matching alone misses.

\subsection{\textsc{CoverageTriple} ($c{=}3.7$)}
\label{app:covtriple}

Decomposes documents at three granularities: sentences, consecutive sentence pairs, and paragraphs. Each granularity produces two aggregation channels, MaxSim for the strongest local match and TopMeanSim for the mean similarity of chunks above the per-query median. A topic-level channel at 128-token truncation and a full-document channel complete the feature set. All channels are debiased by subtracting their respective embedding centroids before similarity computation.

\subsection{\textsc{LexicalHybridRRF} ($c{=}3.9$)}
\label{app:lexhybrid}

Introduces two lexical channels that require no encoder calls: BM25-style IDF-weighted word overlap following the term-weighting principles of \citet{robertson1995okapi}, and word-bigram overlap. These channels are fused with the embedding-based channels through reciprocal rank fusion \citep{cormack2009reciprocal}, $\text{RRF}(d) = \sum_{i} 1/\text{rank}_i(d)$. Both the IDF-weighted lexical matching and the RRF fusion are well-established techniques in information retrieval. The search independently rediscovers the hybrid sparse-dense retrieval paradigm, confirming that exact term correspondences capture signal that continuous embeddings smooth over.

\subsection{\textsc{CrossRoundRRF} ($c{=}3.9$)}
\label{app:crossround}

Introduces iterative query refinement. Its backbone enhances \textsc{LexicalHybridRRF} with two additional lexical channels, query-term coverage ratio and rare-term IDF scoring with Unicode-aware CJK tokenization, for a total of four lexical channels. Round~1 produces an initial ranking via multi-channel RRF. Round~2 applies a Rocchio update, computing positive and negative centroids from documents above and below the per-query median RRF score:
$\mathbf{q}_{\text{rocchio}} = \text{normalize}(\mathbf{q} + \bar{\mathbf{d}}_{\text{pos}} - \bar{\mathbf{d}}_{\text{neg}})$.
This implements classical Rocchio pseudo-relevance feedback \citep{rocchio1971relevance} in the dense embedding space (PRF was a seeded inspiration, Section~\ref{sec:proposer}). Prior dense PRF work includes VPRF \citep{li2023vprf}, which computes a uniform mean over top-$k$ document embeddings and interpolates with the query, and ANCE-PRF \citep{yu2021anceprf}, which trains a neural module to learn the aggregation. CrossRoundRRF applies the full Rocchio formulation with explicit negative-centroid subtraction, adds a query-residual channel, and fuses multiple refinement rounds through RRF rather than linear interpolation. Round~3 computes a query residual $\mathbf{q}_{\text{res}} = \mathbf{q} - \text{proj}(\mathbf{q}, \bar{\mathbf{d}}_{\text{pos}})$ that captures the component of the original query orthogonal to the relevant-document cluster; when this residual is small, $\|\mathbf{q}_{\text{res}}\| < 0.1$, round~3 falls back to round~1. The key design choice is cross-round RRF: rather than taking the element-wise maximum across rounds, the program computes $\text{RRF}(\text{rank}_{R1}, \text{rank}_{R2}, \text{rank}_{R3})$, requiring a document to rank consistently across all rounds.

\subsection{\textsc{DiverseDualCtx} ($c{=}5.6$)}
\label{app:diversedual}

Performs contextual query expansion. Two anchor documents are selected from the top-half of a preliminary ranking: the dominant anchor is the top-ranked document, and the diverse anchor is the top-half document most dissimilar to the dominant anchor by cosine distance. The query is concatenated with each anchor's best-matching sentence and re-encoded under the \texttt{retrieval.query} adapter, producing two new multi-channel scoring rounds. All rounds are fused through cross-round RRF with an adaptive gate.

\subsection{\textsc{ConsensusRocchio} ($c{=}6.4$)}
\label{app:consensus}

Constructs pseudo-relevance feedback centroids using consensus filtering rather than a naive rank cutoff. Pairwise document-document similarities within the top-quarter of the ranking identify a core cluster whose members have above-median mutual similarity. Only core-cluster documents contribute to the positive centroid; outlier top-quarter documents and all bottom-half documents form the negative centroid. This consensus-based selection contrasts with VPRF \citep{li2023vprf}, which takes a uniform mean over all top-$k$ documents without filtering for cluster coherence. The program also performs full bidirectional scoring, encoding queries with the passage adapter and documents with the query adapter, to produce cross-perspective channels.

\subsection{\textsc{NegContrastive} ($c{=}7.2$)}
\label{app:negcontrast}

Constructs a contrastive query embedding by encoding the query concatenated with the best-matching sentence of the bottom-ranked document, then subtracting the result from a positive-anchor expansion: $\mathbf{q}_{\text{contrast}} = \text{normalize}(\mathbf{q}_{\text{pos}} - \mathbf{q}_{\text{neg}})$. This nonlinear contrastive signal captures encoder-level interactions between query and context that linear Rocchio over corpus centroids cannot represent.

\subsection{\textsc{MomentumProg} ($c{=}9.8$)}
\label{app:momentum}

Selects expansion anchors by score momentum rather than absolute rank. The positive anchor is the top-half document whose rank improved most between the baseline and the first refinement round; the negative anchor is the bottom-half document whose rank dropped most. Four expansion rounds each use a different LoRA adapter: \texttt{retrieval.query}, \texttt{text-matching}, \texttt{classification}, and \texttt{retrieval.passage}.

\subsection{\textsc{GraphCentrality} ($c{=}12.2$)}
\label{app:graphcent}

Selects expansion anchors using document-graph centrality. Within the top-quarter of the current ranking, pairwise document-document similarities identify the most central document as the positive anchor. The negative anchor is the bottom-half document closest to the top-quarter centroid, representing the hardest negative.

\subsection{\textsc{FisherStability} ($c{=}14.7$)}
\label{app:fisherstab}

Introduces two zero-cost algebraic scoring channels alongside multi-round contextual expansion. The Fisher discriminant channel computes the direction that maximally separates top-quarter from bottom-quarter documents, $\mathbf{w}_{\text{Fisher}} = \text{normalize}(\bar{\mathbf{d}}_{\text{top}} - \bar{\mathbf{d}}_{\text{bot}})$, and scores each document as $\mathbf{d}_i^\top \mathbf{w}_{\text{Fisher}}$. This is a rediscovery of the Fisher linear discriminant \citep{fisher1936lda}, originally developed for taxonomic classification, here applied as a scoring function in embedding space. The rank-stability channel computes the variance of each document's rank across all preceding scoring channels; low variance indicates robust consensus. Both channels require zero additional encoder calls. The full pipeline aggregates 22 channels through RRF.

\section{Held-Out Transfer and the Learned-Head Baseline}
\label{app:headbaseline}

\begin{figure}[t]
\centering
\includegraphics[width=0.74\columnwidth]{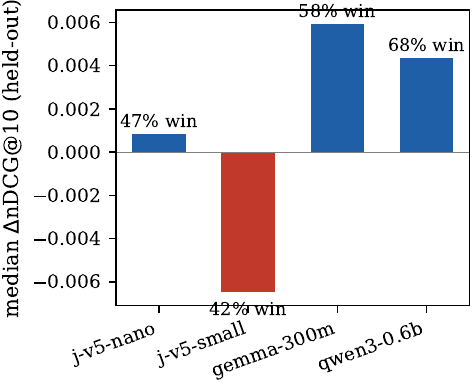}
\caption{Held-out transfer of the representative program \textsc{LexHybridRRF} per encoder family, as median $\Delta$nDCG@10 over the 19 held-out tasks with the win-rate annotated. The gains are positive on the unseen \embgemma{} and \qwenthreeemb{} families and on \jinanano{}, and the program is applied unmodified across all four encoders.}
\label{fig:permodel_bar}
\end{figure}

Figure~\ref{fig:permodel_bar} shows the held-out transfer of the representative program \textsc{LexHybridRRF} per encoder family, with a positive median $\Delta$nDCG@10 and a majority win-rate on the unseen \embgemma{} and \qwenthreeemb{} families and on \jinanano{}, applied unmodified across all four encoders.

The natural training-time alternative to a training-free program keeps the encoder frozen but spends a small amount of training compute on a head and then runs cheap inference at $c{\approx}1$. Because test-time compute is defined by the trade against this training axis \citep{snell2024scalingttc}, we evaluate it directly. We train four head families on the pooled query and positive-document pairs of the 14 discovery tasks, separately per encoder, with in-batch InfoNCE. The families are an unsupervised PCA-whitening map, a $d{\times}d$ linear projection, a residual low-rank metric of rank 64, and a residual two-layer MLP. Each head transforms the frozen embeddings and scores by cosine in the transformed space, with scoring otherwise identical to the baseline. To give the head a fair tuning budget, each family is selected by mean nDCG@10 on a held-out split of discovery queries, then frozen and evaluated both on the discovery tasks and on the 19 held-out tasks. Training is cheap, taking 3 to 4 seconds and 0.1 to 2 M parameters per encoder.

\begin{figure*}[t]
\centering
\includegraphics[width=\textwidth]{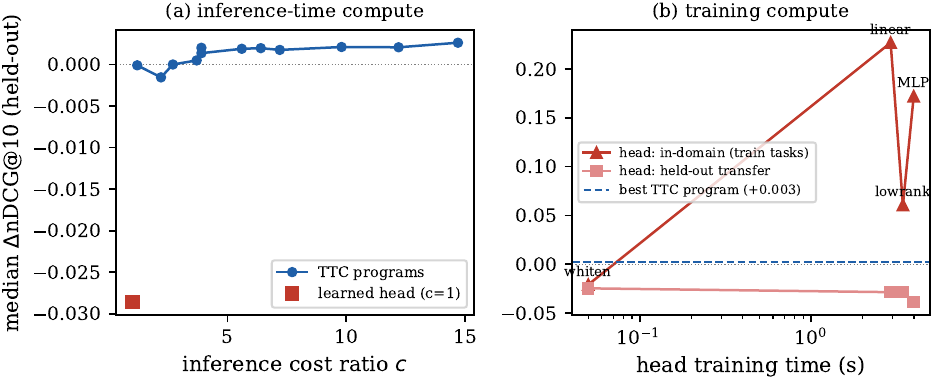}
\caption{Test-time versus training-time compute on the 19 held-out tasks, averaged over the four base encoders. \textbf{(a)}~Inference-time compute, showing the twelve frontier programs against cost ratio $c$ by median held-out $\Delta$nDCG@10, with the learned linear head placed at $c{=}1$ since its inference is a single matmul. \textbf{(b)}~Training-time compute, where learned heads gain on the in-domain training tasks, marked $\triangle$, by up to $+0.23$ but fall below baseline on held-out transfer, marked $\square$, and the dashed line marks the best frontier program's held-out median.}
\label{fig:trainvstest}
\end{figure*}

Figure~\ref{fig:trainvstest} contrasts the two axes. The training-free programs transfer to new encoders and tasks, while a learned head trained at the same data budget does not, even though it fits the discovery tasks far better. Averaged over the four base encoders, the linear head improves discovery-task retrieval by $+0.227$ nDCG@10, ranging from $+0.20$ to $+0.25$ across encoders, and by $+0.044$ on held-out queries of the discovery domains. On the 19 held-out tasks the same head falls by $-0.029$ on average and loses 12 to 14 of 18 tasks per encoder, with $-0.023$ on \jinanano{}, $-0.019$ on \jinasmall{}, $-0.024$ on \embgemma{}, and $-0.049$ on \qwenthreeemb{}. Under a paired bootstrap with $10{,}000$ resamples these held-out regressions are significant on 9 to 13 tasks per encoder, while significant gains number zero to two. The MLP and low-rank heads behave the same way, and unsupervised whitening is mildly negative everywhere. A learned metric memorizes the geometry of the domains it trained on and degrades on unseen ones.

At $c{\ge}4$ the training-free programs instead help the majority of held-out cells, with a positive median $\Delta$nDCG@10 and a per-program win-rate of $54$ to $57\%$ over the pooled (model,~task) cells, and they transfer to encoder families never seen during discovery. Neither axis beats the cosine baseline on the held-out arithmetic mean, as Section~\ref{sec:cost-honest} discusses, so the two differ in which tasks they help rather than in average magnitude. Inference-time structural compute helps the typical new task with no label from it, whereas the learned head helps only tasks drawn from its training distribution, which is why test-time compute rather than a cheap learned head is the axis that transfers with no per-domain labels.

\section{Comparison with Classical Rocchio and PRF Baselines}
\label{app:rocchio-comparison}

Classical Rocchio \citep{rocchio1971relevance} is the strongest training-free dense PRF baseline and is mathematically equivalent to the uniform-mean vector PRF of \citet{li2023vprf}. To position \textsc{SoftCentroid} against this baseline at parity, we grid sixteen classical-Rocchio configurations over $K\in\{2,3,5,10\}$ and $\beta\in\{0.1,0.3,0.5,0.7\}$ across all 13 nanoBEIR tasks on both models, then compare \textsc{SoftCentroid} against the best Rocchio per cell.

\begin{figure}[htb]
\centering
\begin{tikzpicture}
\begin{axis}[
  ybar,
  area legend,
  width=\columnwidth,
  height=4.6cm,
  bar width=6pt,
  ymajorgrids=true,
  major grid style={draw=gray!30},
  ymin=0, ymax=14,
  ylabel={$\Delta$ nDCG@10 over cosine},
  ylabel style={font=\scriptsize},
  symbolic x coords={e5-small, e5-base, e5-large, gte-base, bge-large, jina-small, jina-nano},
  xtick=data,
  x tick label style={font=\scriptsize, rotate=30, anchor=east},
  yticklabel style={font=\scriptsize},
  enlarge x limits=0.08,
  legend style={font=\scriptsize, at={(0.5,1.02)}, anchor=south, legend columns=2, draw=none},
  axis line style={draw=gray!50},
]
\addplot[fill=gray!50, draw=gray!70] coordinates {
  (e5-small, 5.68)  (e5-base, 6.55)  (e5-large, 4.99)
  (gte-base, 1.25)  (bge-large, 1.13) (jina-small, 0.93) (jina-nano, 1.22)};
\addplot[fill=blue!55, draw=blue!75] coordinates {
  (e5-small, 9.47)  (e5-base, 12.55) (e5-large, 10.27)
  (gte-base, 3.56)  (bge-large, 5.03) (jina-small, 4.74) (jina-nano, 5.64)};
\legend{best Rocchio, \textsc{SoftCentroid} default}
\end{axis}
\end{tikzpicture}
\caption{ArguAna full-BEIR lift on each of the seven embedding-model families. \textsc{SoftCentroid} at the universal default $K{=}3, \alpha{=}0.5, \tau{=}0.05$ exceeds the best classical Rocchio configuration on every model. Every bar is statistically significant at $p<10^{-4}$ on a paired bootstrap with $10{,}000$ resamples.}
\label{fig:rocchio-bar}
\end{figure}
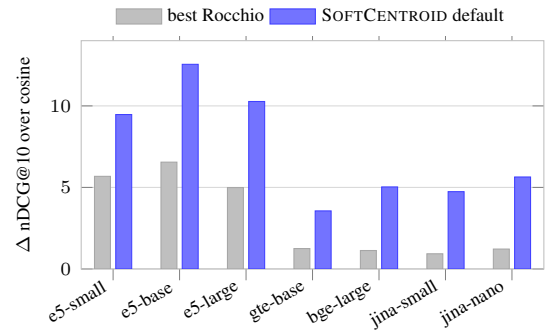

Table~\ref{tab:rocchio-vs-sc} and Figure~\ref{fig:rocchio-bar} report the comparison on full-BEIR ArguAna across all seven embedding-model families. The softmax-weighted \textsc{SoftCentroid} default exceeds the best Rocchio configuration on every model by $+2.31$ to $+6.00$ nDCG@10, every cell at $p<10^{-4}$. On NFCorpus, SciFact, and FiQA-2018 the two methods land within one nDCG@10 point. The separation concentrates on symmetric retrieval, where the cosine geometry favours centroid replacement.

\begin{table}[htb]
\centering
\small
\setlength{\tabcolsep}{2pt}
\begin{tabular}{@{}lrrr@{}}
\toprule
Encoder & best Rocchio & \textsc{SC} & \textsc{SC} advantage \\
\midrule
\efivesmall{}  & $+5.68$ & $+9.47$  & $+3.79$ \\
\efivebase{}   & $+6.55$ & $+12.55$ & $+6.00$ \\
\efivelarge{}  & $+4.99$ & $+10.27$ & $+5.28$ \\
\gtebase{}     & $+1.25$ & $+3.56$  & $+2.31$ \\
\bgelarge{}    & $+1.13$ & $+5.03$  & $+3.90$ \\
\jinasmall{}   & $+0.93$ & $+4.74$  & $+3.81$ \\
\jinanano{}    & $+1.22$ & $+5.64$  & $+4.42$ \\
\bottomrule
\end{tabular}
\caption{\textsc{SoftCentroid} versus best classical Rocchio on full-BEIR ArguAna across all seven embedding-model families. Each value is $\Delta$ nDCG@10 ($\times 100$) over the cosine baseline. The Rocchio number is the best of a sixteen-cell $(K, \beta)$ grid. \textsc{SC} uses the universal default $K{=}3, \alpha{=}0.5, \tau{=}0.05$. All $14$ method-model cells are statistically significant at $p<10^{-4}$ on a paired bootstrap with $10{,}000$ resamples. The softmax-weighted variant exceeds the uniform-mean baseline on every model by $+2.31$ to $+6.00$ nDCG@10.}
\label{tab:rocchio-vs-sc}
\end{table}

\section{Baseline Reproduction}
\label{app:baseline}

\begin{table}[htb]
\centering
\small
\setlength{\tabcolsep}{5pt}
\begin{tabular}{lccrr}
\toprule
Model & Task & Pub. & Ours & $\Delta$ \\
\midrule
\multirow{4}{*}{\jinasmall{}} & NFCorpus & 39.81 & 39.76 & $-0.05$ \\
                          & SciFact  & 76.53 & 76.59 & $+0.06$ \\
                          & ArguAna  & 65.07 & 64.71 & $-0.36$ \\
                          & FiQA-2018 & 49.63 & 49.52 & $-0.11$ \\
\midrule
\multirow{4}{*}{\jinanano{}}  & NFCorpus & 38.69 & 38.75 & $+0.07$ \\
                          & SciFact  & 75.78 & 75.91 & $+0.13$ \\
                          & ArguAna  & 65.70 & 65.63 & $-0.07$ \\
                          & FiQA-2018 & 47.85 & 47.87 & $+0.02$ \\
\bottomrule
\end{tabular}
\caption{Reproduction of the trivial program $P_0$, the cosine baseline with single forward pass, reporting nDCG@10 $\times 100$ for \jinasmall{} and \jinanano{} on four BEIR tasks. ``Pub.'' is the published leaderboard number and ``Ours'' is our fp16 reproduction. All eight cells lie within 0.4 nDCG points.}
\label{tab:baseline}
\end{table}

\end{document}